\documentclass[twocolumn, twoside, 10pt, letterpaper, table]{scrartcl}

\usepackage[utf8]{inputenc}
\usepackage[T1]{fontenc}

\usepackage[
    margin=1.85cm,
    footskip=1cm,
  ]{geometry}

\usepackage{times}
\addtokomafont{sectioning}{\rmfamily}

\RedeclareSectionCommands[
    beforeskip=-.5\baselineskip,
    afterskip=.25\baselineskip
  ]{section,subsection,subsubsection}
\RedeclareSectionCommands[
    beforeskip=.5\baselineskip,
    afterskip=-1em
  ]{paragraph,subparagraph}
\usepackage{abstract}

\usepackage{microtype}
\usepackage{siunitx}
\DeclareSIUnit\px{px}
\usepackage[super]{nth}

\usepackage{amsmath}
\usepackage{amsfonts}
\usepackage{amssymb}
\usepackage{amsthm}
\usepackage{mleftright}
\providecommand{\mathsymbol}[1]{\ensuremath{\boldsymbol{#1}}}
\let\temp\rmdefault
\usepackage{mathpazo}
\let\rmdefault\temp
\usepackage{upgreek}
\newcommand\Tau{\mathrm{T}}
\DeclareMathAlphabet{\mathbit}{OT1}{cmr}{bx}{it}
\usepackage{xstring}
\newcommand{\function}[1][f]{\ensuremath{\operatorname{#1}}}
\newcommand{\functionAt}[2][f]{\ensuremath{\ifthenelse{\equal{#1}{}}{}{\function[#1]}\mleft(#2\mright)}}
\newcommand{\V}[1]{\ensuremath{\mathsymbol{\lowercase{#1}}}}
\newcommand{\VecDef}[1]{\mleft( #1 \mright)}
\newcommand{\Set}[1]{\ensuremath{\mathcal{\uppercase{#1}}}}
\newcommand{\datal}{x}
\newcommand{\datalabel}{\ensuremath{y}}
\newcommand{\DataVec}{\ensuremath{\V{\datal}}}
\newcommand{\LabelVec}{\ensuremath{\V{\datalabel}}}
\newcommand{\abs}[1]{\ensuremath{\left\lvert #1 \right\rvert}}
\renewcommand{\Tau}{\Set{T}}

\usepackage[autostyle=true]{csquotes}

\usepackage[inline]{enumitem}

\usepackage[numbers, sort]{natbib}

\usepackage{glossaries}
\newacronym{fps}{FPS}{frames per second}
\newacronym{rsu}{RSU}{road scene understanding}
\newacronym{gta}{GTA~V}{Grand Theft Auto V}
\newacronym{side}{SIDE}{single-image depth estimation}
\newacronym{fcn}{FCN}{fully-convolutional network}
\newacronym{cnn}{CNN}{convolutional neural network}
\newacronym{miou}{MIoU}{mean intersection over union}
\newacronym{relu}{ReLU}{rectified linear unit}
\newacronym{aspp}{ASPP}{atrous spatial pyramid pooling}
\newacronym{resnet}{ResNet}{residual network}
\newacronym{rmsctd}{RMSCTD}{root mean squared cyclic time difference}
\newacronym{mse}{MSE}{mean squared error}
\newacronym{rmse}{RMSE}{root mean squared error}
\newacronym[shortplural=ADAS]{adas}{ADAS}{advanced driver assistance system}
\newglossaryentry{cv}{name=computer vision, description={}}
\newglossaryentry{av}{name=autonomous vehicles, description={}}
\newglossaryentry{ad}{name=autonomous driving, description={}}
\newglossaryentry{mt}{name=multi-task, description={}}
\newglossaryentry{si}{name=synthetic images, description={}}
\newglossaryentry{ann}{name=artificial neural network, description={}}
\newglossaryentry{cg}{name=computer graphics, description={}}
\newglossaryentry{std}{name=synthetic training data, description={}}
\newglossaryentry{aux}{name=auxiliary task, description={}}
\newglossaryentry{main}{name=main task, description={}}
\newglossaryentry{semseg}{name=semantic segmentation, description={}}
\newglossaryentry{ir}{name=image recognition, description={}}
\newglossaryentry{ac}{name=atrous convolution, description={}}
\newglossaryentry{conv}{name=convolutional, description={}}
\newglossaryentry{max}{name=max pooling, description={}}
\newglossaryentry{fc}{name=fully connected, description={}}
\newglossaryentry{softmax}{name=softmax, description={}}
\newglossaryentry{ce}{name=cross entropy, description={}}
\newglossaryentry{batch}{name=mini batch, description={}}
\newglossaryentry{ow}{name=open world, description={}}
\newglossaryentry{od}{name=object detection, description={}}
\newglossaryentry{sctd}{name=squared cyclic time difference, description={}}
\newglossaryentry{loss}{name=loss function, description={}}
\newglossaryentry{ic}{name=image classification, description={}}
\newglossaryentry{at}{name=atomic task, description={}}
\newglossaryentry{vb}{name=vision-based, description={}}
\newglossaryentry{e-d}{name=encoder-decoder, description={}}
\newglossaryentry{e}{name=encoder, description={}}
\newglossaryentry{d}{name=decoder, description={}}
\newglossaryentry{dataset}{name=synMT, description={}}

\glsdisablehyper  

\newcommand{\ia}{\textit{i{.}a{.}}, }
\newcommand{\ie}{\textit{i{.}e{.}}, }
\newcommand{\eg}{\textit{e{.}g{.}}, }

\usepackage{graphicx}
\usepackage{tikz}
\usepackage{xcolor}
\definecolor{tumblue}{RGB}{0,101,189}
\definecolor{tumbluelight}{RGB}{152,198,234}

\usepackage{tabularx}
\newcolumntype{Y}{>{\raggedleft\arraybackslash}X}
\usepackage{multicol}
\newcommand{\header}[2]{\multicolumn{1}{#1}{\footnotesize #2}}

\newcommand{\lefthead}[1]{{\footnotesize #1}}
\newcommand{\tabsuffix}[1]{\parbox{0.8cm}{\raggedleft\footnotesize (#1)}}

\newcommand{\best}[1]{\cellcolor{tumbluelight} #1}
\usepackage{booktabs}

\usepackage{subcaption}
\setkomafont{caption}{\footnotesize}
\setkomafont{captionlabel}{\usekomafont{caption}\bfseries}

\hypersetup{hidelinks=true}

\usepackage[
    capitalise,
    noabbrev,
  ]{cleveref}

\begin{document}

\title{
  \normalfont
  Auxiliary Tasks in Multi-task Learning
  }

\setkomafont{author}{\large}
\author{
  Lukas Liebel \and
  Marco Körner
  }
\publishers{\normalsize
Computer Vision Research Group, Chair of Remote Sensing Technology \\ Technical University of Munich, Germany \\ \{lukas.liebel, marco.koerner\}@tum.de \\[0.25cm]
}
\date{}

\twocolumn[
  \maketitle
  \begin{onecolabstract}

    \Gls{mt} \glspl{cnn} have shown impressive results for certain combinations of tasks, such as \gls{side} and \gls{semseg}.
    This is achieved by pushing the network towards learning a robust representation that generalizes well to different \glspl{at}.
    We extend this concept by adding \glspl{aux}, which are of minor relevance for the application, to the set of learned tasks.
    As a kind of additional regularization, they are expected to boost the performance of the ultimately desired \glspl{main}.
    To study the proposed approach, we picked \gls{vb} \gls{rsu} as an exemplary application.
    Since \gls{mt} learning requires specialized datasets, particularly when using extensive sets of tasks, we provide a multi-modal dataset for \gls{mt} \gls{rsu}, called \gls{dataset}.
    More than \num{2.5e5} synthetic images, annotated with \num{21} different labels, were acquired from the video game \gls{gta}.
    Our proposed deep \gls{mt} \gls{cnn} architecture was trained on various combination of tasks using \gls{dataset}.
    The experiments confirmed that \glspl{aux} can indeed boost network performance, both in terms of final results and training time.

  \end{onecolabstract}
  \vspace{1cm}
]

\glsresetall
\defglsentryfmt{\ifglsused{\glslabel}{\glsgenentryfmt}{\emph{\glsgenentryfmt}}}


\section{Introduction}
\label{sec:introduction}

Various applications require solving several \glspl{at} from the \gls{cv} domain using a single image as input.
Such basic tasks often include \gls{side} \citep{Eigen14, Eigen15, Laina16, Liu15}, \gls{semseg} \citep{Chen17, Chen18, Long15, Badrinarayanan17}, or \gls{ic} \citep{He16, Szegedy16, Simonyan14}.
While each task is traditionally tackled individually, close connections between them exist.
Exploiting those by solving them jointly can increase the performance of each individual task and save time during training and inference.
\Gls{mt} learning, \ie the concept of learning several outputs from a single input simultaneously, was applied to numerous tasks and techniques, including \gls{ann} architectures with hard parameter sharing, within the last decades \citep{Caruana98,Caruana93,Ruder17}.
\Gls{cnn} architectures for \gls{mt} learning were proven successful, especially for the combination of \gls{od} and \gls{semseg} \citep{Girshick15, Yao12}, as well as \gls{side} and \gls{semseg} \citep{Eigen15, Kendall18}.
In order to combine the contributing single-task \glspl{loss} to a final \gls{mt} loss, subject to optimization, \citet{Kendall18} very recently proposed an approach for learning a weighting between them, in order to address their unequal properties and behaviors.

Building upon these concepts, we introduce \glspl{aux} to \gls{mt} setups.
Apart from the \glspl{main}, which are the ultimately required output for an application, \glspl{aux} serve solely for learning a rich and robust common representation of an image.
By choosing tasks that are easy to learn and support the \glspl{main}, this helps to boost the system during the training stage.
A similar concept was proposed by \citet{Romera12}.
Contrary to their notion of \emph{principal} and \gls{aux}, we define \glspl{aux} as \emph{seemingly} unrelated tasks that in fact base on similar features as the \glspl{main}.
We, therefore, neither explicitly model them as different groups, nor impose any penalties to encourage orthogonal representations.

\begin{figure}[t]
  \centering

  \includegraphics{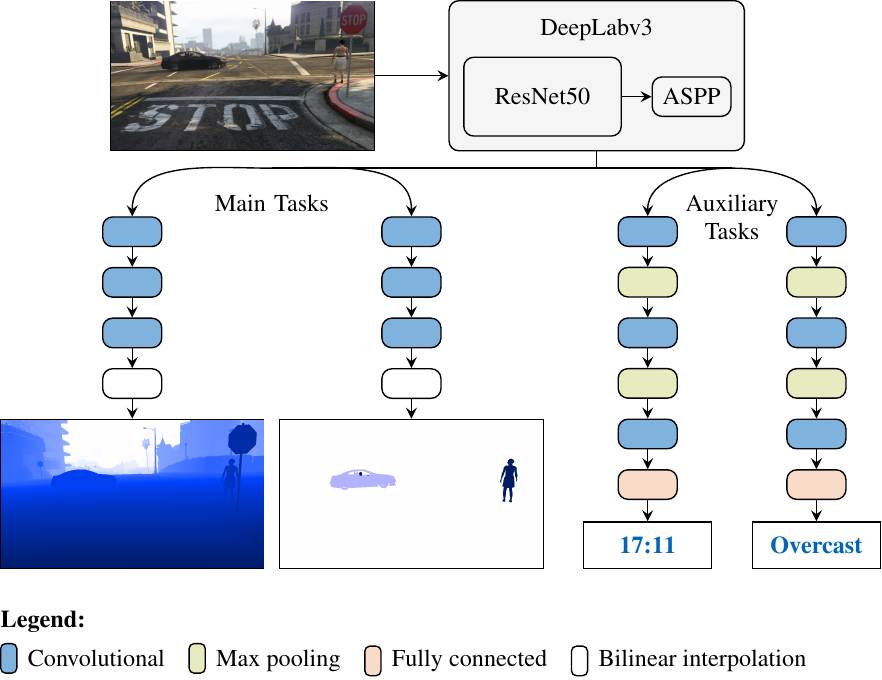}

  \caption{Schematic overview of a first implementation of the proposed concept for the utilization of \glspl{aux} in \gls{mt} learning applied to \gls{vb} \gls{rsu}.}
  \label{fig:overview}

\end{figure}

In order to analyze our concept, we have chosen \gls{vb} \gls{rsu} as an exemplary application.
\Glspl{adas} and \gls{av} need to gather information about the surrounding of the vehicle, in order to be able to safely guide the driver or vehicle itself through complex traffic scenes.
Apart from traffic signs and lane markings, typical components of such scenes that need to be considered are other road users, \ie primarily vehicles and pedestrians.
When it comes to decisionmaking, additional important local or global parameters will certainly be taken into account.
Those include, \ia object distances and positions, or the current time of day \citep{Ma16} and weather conditions \citep{Yan09,Roser08}.
The application of \gls{vb} methodology seems natural in the context of \gls{rsu}, since the rules of the road and signage were designed for humans who mainly rely on visual inspection in this regard.
\Cref{fig:overview} shows an overview of our network architecture and illustrates the concept of \glspl{aux}, with \gls{side} and \gls{semseg} serving as \glspl{main} and the estimation of the time of day and weather conditions as \glspl{aux}.

Deep learning techniques, which represent the state of the art in virtually all of the aforementioned sub-tasks of \gls{rsu}, heavily rely on the availability of large amounts of training data.
The availability of specialized datasets, such as KITTI \citep{Geiger13} and Cityscapes \citep{Cordts16}, facilitated \gls{cnn}-based visual \gls{rsu}.
For \gls{mt} learning, annotated datasets with various labels are required.
Manual annotation of training data is, however, a time-consuming and hence expensive task.
Moreover, it is hard to ensure that all possibly occurring phenomena and their numerous combinations are covered by the training data.
Some phenomena cannot be covered by real data at all.
Typical example for such scenes in the context of \gls{ad} are traffic accidents, which are of great interest for solving this task but luckily happen rarely.
Intentionally causing those situations is often only possible at great expenses or even outright impossible.

As the development of techniques for creating near photo-realistic \gls{si} in the field of \gls{cg} rapidly advances, a growing interest in \gls{std} arose in recent years.
Simulations allow for the acquisition of virtually unlimited amounts of automatically labeled training data with relatively little effort.
In their pioneering work, \citet{Richter16} proposed a method for the propagation of manual annotations to following frames of a sequence using additional information extracted from intermediate rendering steps of the popular video game \gls{gta}.
Following publications proposed methods for the fully-automatic generation of training data from video games \citep{Richter17, Johnson-Roberson17} and custom simulators based on game engines \citep{Mueller18} or procedural models \citep{Tsirikoglou17}.
Widespread synthetic datasets for visual \gls{rsu} include Virtual KITTI \citep{Gaidon16}, SYNTHIA \citep{Ros16}, and CARLA \citep{Dosovitskiy17}.

Based on the current state of the art, we discuss the concept of introducing \glspl{aux} to \gls{mt} setups in the remainder of this paper.
Our main contributions are:
\begin{enumerate*}[label=\roman*.)]
  \item A novel concept for boosting training and final performance of \gls{mt} \glspl{cnn} by utilizing seemingly unrelated \glspl{aux} for regularization,
  \item a first implementation and study of this concept in the context of \gls{vb} \gls{mt} \gls{rsu},
  \item a synthetic dataset, called \gls{dataset}, for \gls{rsu} containing a multitude of multi-modal ground-truth labels\footnote{The synMT dataset and source code utilized to acquire it are publicly available at \url{http://www.lmf.bgu.tum.de/en/synmt}}, and
  \item a \gls{mt} \gls{cnn} architecture building upon the work of \citet{Chen17} and \citet{Kendall18}.
\end{enumerate*}

\section{Introducing Auxiliary Tasks to Multi-task Learning}
\label{sec:approach}

The general idea of \gls{mt} learning is to find a common representation in the earlier layers of the network, while the individual tasks $\tau \in \Set{\Tau}$ are solved in their respective single-task branches in the later stages of the network.
This is most commonly realized as an \gls{e-d} structure, in which each \gls{at} represents a specialized \gls{d} to the representation provided by the common \gls{e}.
While each type of label $\V{y}_\Set{\Tau} = \VecDef{y_{\tau_{1}}, y_{\tau_{2}}, \ldots} \in \Set{y_\Set{\Tau}}$ favors the learning of certain features in the common part, some of them can be exploited by other tasks as well.
This structure can thus help to boost the performance of the \glspl{at}.

We extend this concept by adding \glspl{aux} to \Set{\Tau}.
We refer to \glspl{aux} as \glspl{at} that are of minor interest or even irrelevant for the application.
Despite being seemingly unrelated, they are expected to assist in finding a rich and robust representation of the input data $\DataVec$ in the common part, from which the ultimately desired \glspl{main} profit.
\Glspl{aux} should be chosen, such that they are easy to be learned and use labels that can be obtained with low effort, \eg global descriptions of the scene.
By forcing the network to generalize to even more tasks, learning \glspl{aux} restricts the parameter space during optimization.
Since, by design, \glspl{aux} are simple, robust, and uncorrelated with the \glspl{main} to a certain extent, they can thus be regarded as a regularizer.

In order to be able to optimize a \gls{mt} network for the learnable parameters $\V{\omega}_\Set{\Tau} = \VecDef{\V{\theta}_\Set{\Tau}}$, which in this case only consist of the network parameters $\V{\theta}_\Set{\Tau}$, a \gls{mt} \gls{loss} $\functionAt[L_\Set{\Tau}]{\DataVec, \LabelVec_\Set{\Tau}, \LabelVec_\Set{\Tau}^\prime; \V{\omega}_\Set{\Tau}}$, comparing ground-truth labels $\V{y}_\Set{\Tau}$ to predictions $\V{y}_\Set{\Tau}^\prime$, has to be designed.
The final \gls{mt} loss, subject to optimization, is a combination of the single-task losses.
Since each of the contributing single-task loss functions $\functionAt[L_\text{$\tau$}]{\DataVec, \LabelVec_\tau, \LabelVec_\tau^\prime; \V{\omega}_\tau}$ may behave differently, weighting each with a factor $c_\tau$ is essential.
This yields a combined \gls{loss}
\begin{align}
  \functionAt[L_\text{comb}]{\DataVec, \LabelVec_\Set{\Tau}, \LabelVec_\Set{\Tau}^\prime ; \V{\omega}_\Set{\Tau}} &=
  \sum_{\tau \in \Set{\Tau}} \functionAt[L_\text{$\tau$}]{\DataVec, \datalabel_\tau, \datalabel_\tau^\prime ; \V{\omega}_\tau} \cdot c_\tau \quad \text{.}
\end{align}
Instead of manually tuning $c_\tau$ to account for the differing variances and offsets amongst the single-task losses, the coefficients can be added to the learnable network parameters $\V{\omega}_\Set{\Tau} = \VecDef{\V{\theta}_\Set{\Tau}, \V{c}_\Set{\Tau}}$.
This, however, requires augmenting \function[L_\text{comb}] with a regularization term \functionAt[R]{c_\tau} to avoid trivial solutions.
We adapt the regularization term $\functionAt[R_\text{log}]{c_\tau} = \log \left ( c_\tau^2 \right )$ suggested by \citet{Kendall18} slightly to
$\functionAt[R_\text{pos}]{c_\tau} = \ln \left ( 1 + c_\tau^2 \right )$ in order to enforce positive regularization values.
Thus, decreasing $c_\tau$ to $c_\tau^2 < 1$ no longer yields negative loss values.
We consider these considerations in our final \gls{mt} loss
\begin{align}
  \begin{split}
    \functionAt[L_\text{$\Set{\Tau}$}]{\DataVec, \LabelVec_\Set{\Tau}, \LabelVec_\Set{\Tau}^\prime ; \V{\omega}_\Set{\Tau}} =
    &\sum_{\tau \in \Set{\Tau}} \frac{1}{2 \cdot c_\tau^2} \cdot
    \functionAt[L_\text{$\tau$}]{\DataVec, \datalabel_\tau, \datalabel_\tau^\prime ; \V{\omega}_\tau} \\
    &+ \functionAt[\ln]{ 1 + c_\tau^2 } \quad \text{.}
  \end{split}
\end{align}

\section{Experiments and Results}
\label{sec:results}

For an experimental evaluation of the approach presented in \cref{sec:approach}, we chose \gls{vb} \gls{rsu} as an exemplary application.
In \gls{rsu}, various tasks have to be solved simultaneously, making it a prime example for \gls{mt} approaches.
\Gls{mt} training requires several different annotations for each image, especially when additional labels for \glspl{aux} are needed, and an overall large number of samples.
Since no suitable dataset was available, we compiled a dataset, called \gls{dataset}, for our experiments ourselves.
This dataset will be made available to the public and is thus described in detail in the following section.
Utilizing \gls{dataset}, we conducted experiments on \gls{mt} training with \glspl{aux}.
A description of the used network architecture and the conducted experiments is given in \cref{subsec:training}, followed by a discussion of the results in \cref{subsec:results}.

\subsection{A Synthetic Dataset for Multi-task Road Scene Understanding}
\label{subsec:data}

Datasets for \gls{rsu}, such as KITTI \citep{Geiger13} and Cityscapes \citep{Cordts16} exist but provide few annotations per image.
Often, manual annotation is not feasible without major effort.
In contrast to this, the simulation of traffic scenes enables fully-automatic labeling but is a sophisticated task which requires extensive knowledge from the field of \gls{cg}.
The video game industry invests heavily in creating photorealistic renderings and dynamic simulations, especially for games from the \gls{ow} genre, which allow the player to freely explore a virtual world.
Making use of techniques developed in this field, synthetic datasets for \gls{rsu} have been presented \citep{Ros16,Dosovitskiy17,Richter16,Richter17,Gaidon16}.
Not only does simulation allow to freely set all parameters of the scene, it also facilitates the creation of error-free labels.

\begin{figure}

  \begin{subfigure}{\linewidth}
    \begin{tikzpicture}
      \node[anchor=south west,inner sep=0] at (0,0) {\includegraphics[height=4.25cm, trim=115 0 115 25, clip]{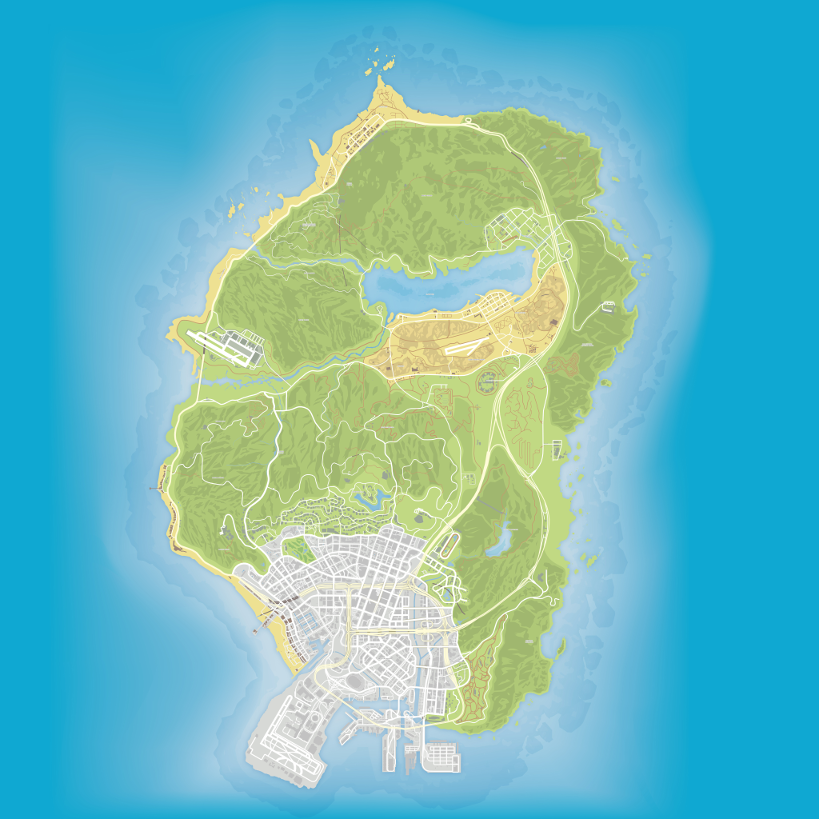}};
      \draw[draw=tumblue, thick] (0.984cm,0.789cm) rectangle (1.821cm,1.451cm);
    \end{tikzpicture}
    \hfill
    \includegraphics[height=4.25cm]{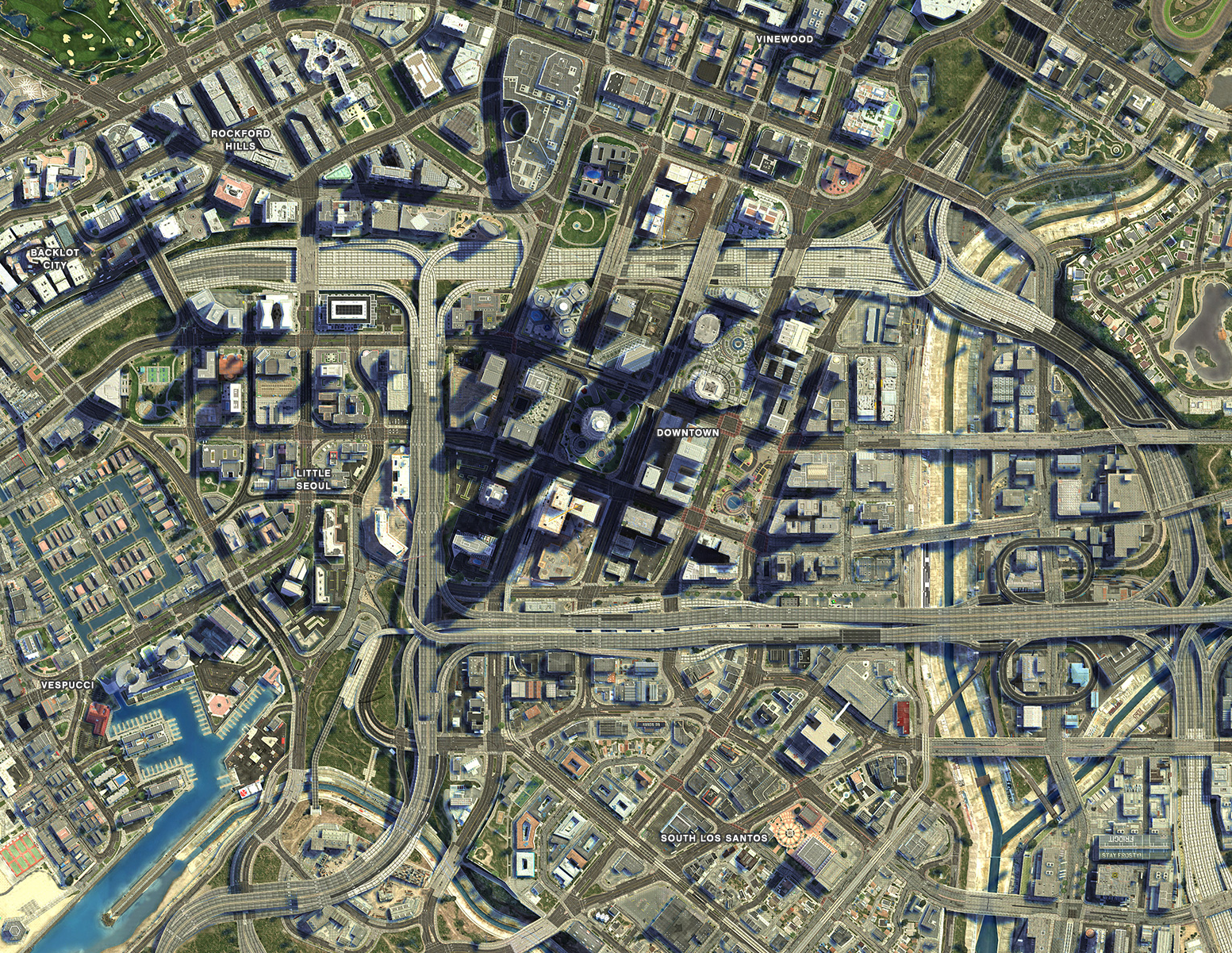}
    \subcaption{GTA V (Images: Rockstar Games)}
  \end{subfigure}

  \begin{subfigure}{\linewidth}
    \centering
    \includegraphics[width=\linewidth, trim=0 0 0 0, clip]{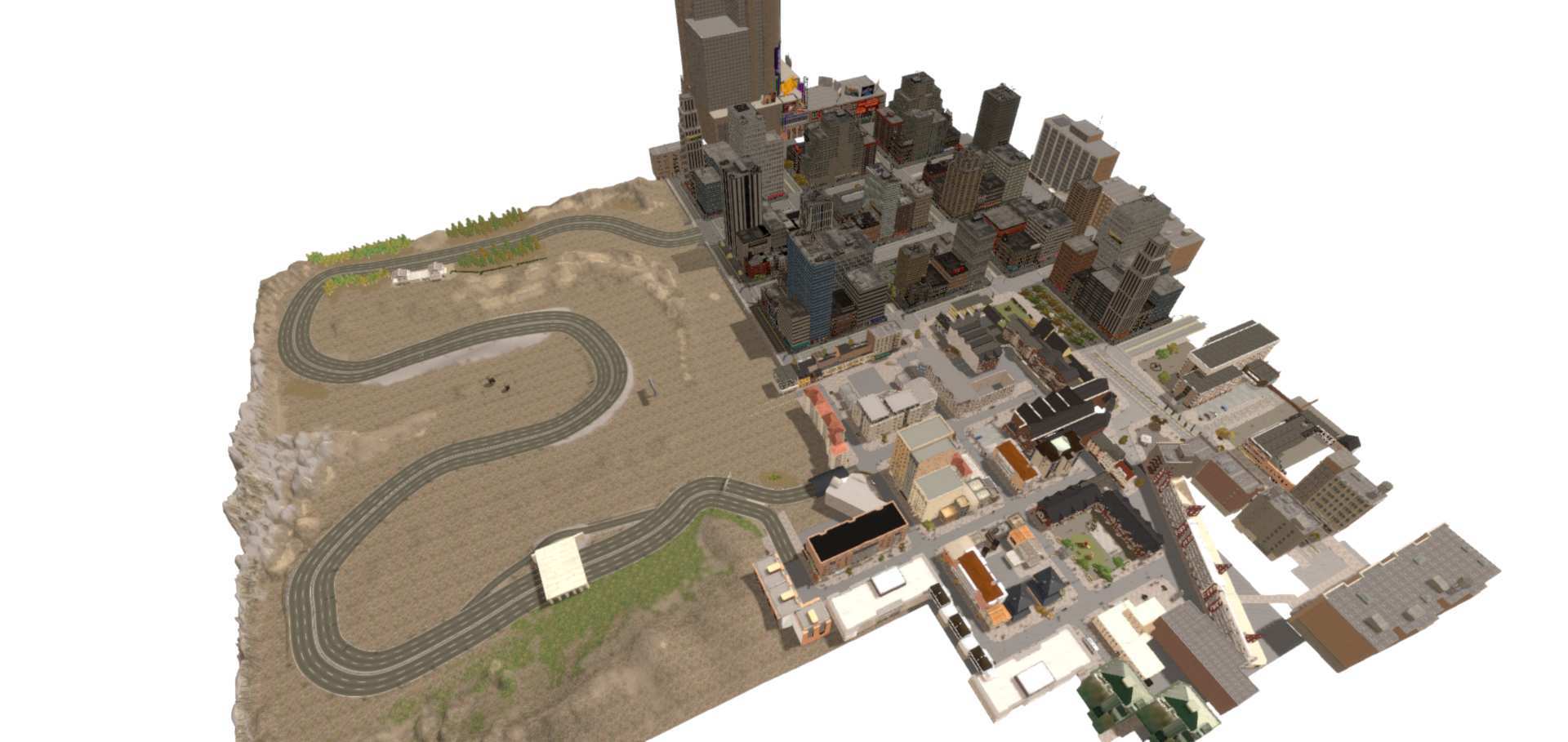}
    \subcaption{SYNTHIA (Image: \citet{Ros16})}
  \end{subfigure}

  \caption{(a) The accessible area in \gls{gta} with a map of the whole island (left) and a detail of the main city, given as a simulated satellite image (right) in comparison to (b) the city area of SYNTHIA \citep{Ros16}.}
  \label{fig:maps}

\end{figure}

\Gls{gta} is the latest title in a series of \gls{ow} video games that provide the player with a realistic environment set in modern-time USA.
We decided upon using \gls{gta} as the source for our dataset, primarily because of its state-of-the-art graphics and the availability of prior work regarding the extraction of annotations.
The game is set on a large island with urban areas, countryside, and highways.
\Cref{fig:maps} shows a map of the accessible area of \gls{gta}, which outranks other simulation sceneries, such as the virtual city model of SYNTHIA \citep{Ros16}, by far.
The player is able to take part in the simulated traffic with various vehicles.

In order to acquire data in the most realistic way possible, we set up a simulator consisting of a gaming PC and likewise periphery.
The computer features an NVIDIA GeForce GTX 1080 GPU, complemented by an Intel i7-7700 CPU and a PCIe NVMe SSD.
This allows running state-of-the-art video games in high resolution and quality while recording data with a high framerate simultaneously.
To enhance the driving experience and thus enable our drivers to steer their car as naturalistic as possible, we equipped our machine with a Logitech G920 force-feedback steering wheel and pedals.
We instructed the drivers to obey the rules of the road and adapt to the simulated traffic.

\begin{table*}

  \caption{Contents of the proposed synthetic dataset \gls{dataset} for \gls{mt} \gls{rsu}, acquired in the video game \gls{gta}, illustrated by a single example. The dataset consists of more than \num{2.5e5} annotated images with a resolution of \SI{1028x720}{\px}, sampled at \SI{15}{fps}.}
  \label{tab:labels}

  \centering

  \begin{subfigure}{.32\linewidth}
    \fbox{\includegraphics[width=\linewidth]{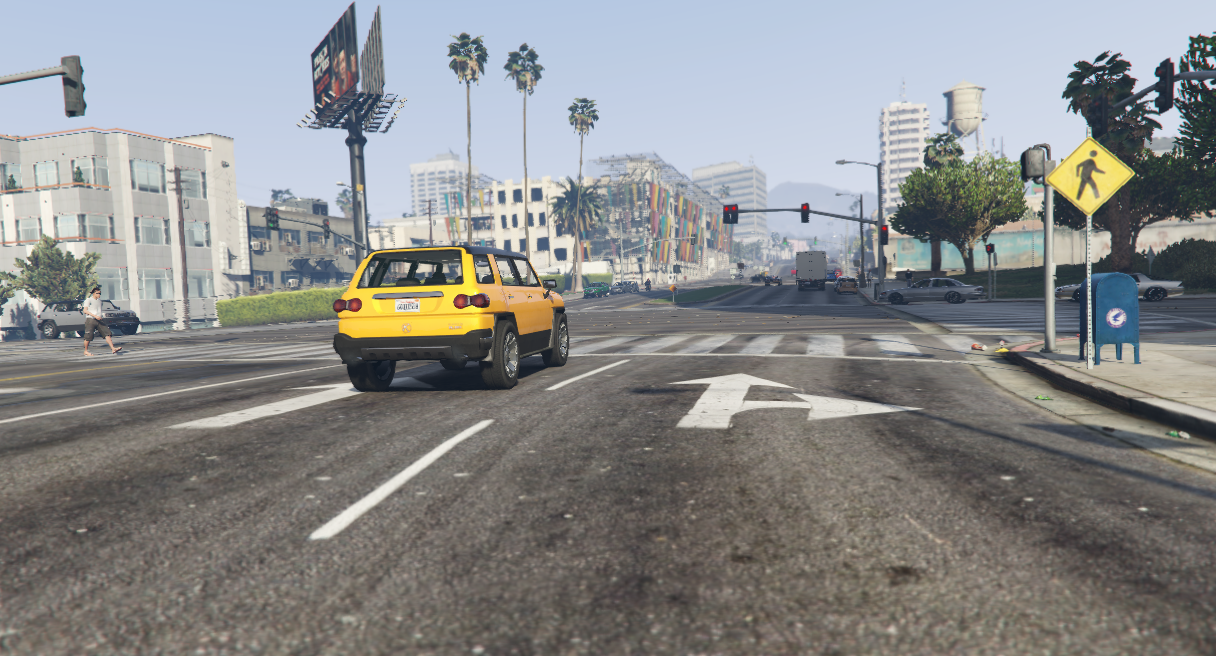}}
    \subcaption{RGB}
  \end{subfigure}
  \hfill
  \begin{subfigure}{.32\linewidth}
    \fbox{\includegraphics[width=\linewidth]{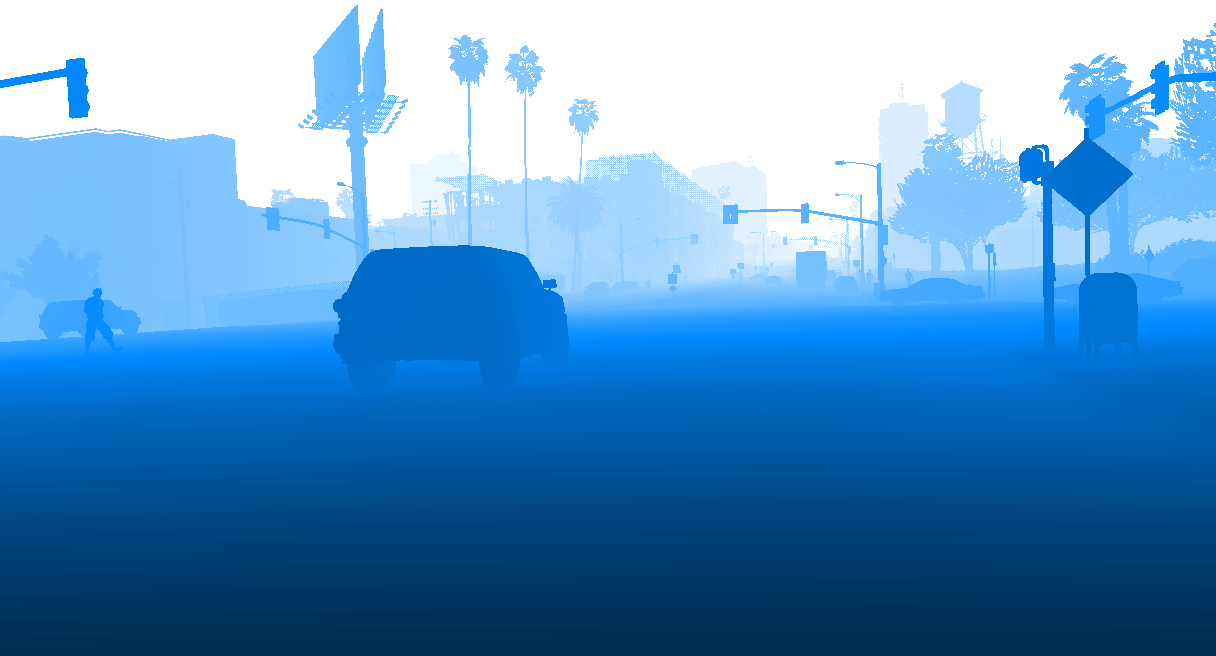}}
    \subcaption{Depth Map}
  \end{subfigure}
  \hfill
  \begin{subfigure}{.32\linewidth}
    \fbox{\includegraphics[width=\linewidth]{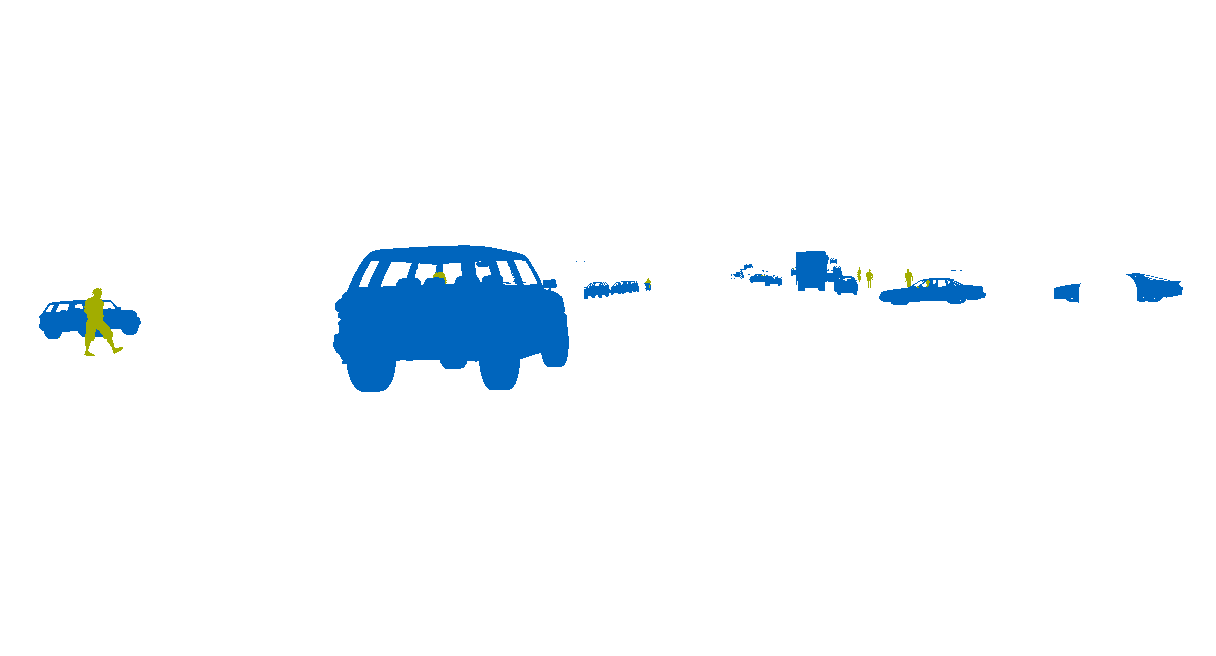}}
    \subcaption{Semantic Masks}
  \end{subfigure}

  \begin{subtable}{\linewidth}
    \begin{tabularx}{\linewidth}{@{}*{2}{XY@{\hspace{1.25cm}}}XY@{}}
      \toprule
      \lefthead{Vehicle:}   & Car                       & \lefthead{Time of Day:}     & 09:11:21                           & \lefthead{Steering:}        & \SI{-67.5}{\degree}               \\
      \lefthead{Model:}     & Asea                      & \lefthead{Weather:}         & Clear                             & \lefthead{Throttle:}        & $0.08$                              \\
      \lefthead{Location:}  & $-1049.59$ \tabsuffix{N}  & \lefthead{Rain Level:}      & $0$                                 & \lefthead{Brake:}           & $0$                                 \\
                            &  $-779.87$ \tabsuffix{E}  & \lefthead{Snow Level:}      & $0$                                 & \lefthead{Gear:}            & \nth{2}                           \\
                            &    $18.76$ \tabsuffix{h}  & \lefthead{Wind Speed:}      & \SI{16.95}{\kilo\meter\per\hour}  & \lefthead{Speed:}           & \SI{24.16}{\kilo\meter\per\hour}  \\
      \lefthead{Heading:}   & $2.34$ \tabsuffix{Pitch}  & \lefthead{Wind Direction:}  & $0.0$ \tabsuffix{N}             & \lefthead{Velocity Vector:} & $-5.35$ \tabsuffix{Pitch}     \\
                            & $-1.21$ \tabsuffix{Roll}  &                             & $-1.0$ \tabsuffix{E}            &                             & $4.06$ \tabsuffix{Roll}       \\
                            & $54.55$ \tabsuffix{Yaw} &                               & $0.0$ \tabsuffix{h}             &                             & $2.62$ \tabsuffix{Yaw}        \\
      \addlinespace
      \lefthead{Past Events:} & \multicolumn{5}{l@{}}{\SI{315}{\second} ago {\footnotesize (Driven in Wrong Lane)}, \enskip \SI{10815.02}{\second} ago {\footnotesize (Crash, Car)}, \enskip \SI{89018.60}{\second} ago {\footnotesize (Crash, Pedestrian)}}\\
      \bottomrule
    \end{tabularx}
    \subcaption{Other Labels}
  \end{subtable}

\end{table*}

The acquired ground-truth labels can be divided into two groups according to the used method for acquisition.
Pixel-wise depth maps and semantic masks, as required for tasks such as \gls{side} and \gls{semseg}, were extracted from the respective buffers using DirectX functions.
Labels for all other tasks were obtained by utilizing the Script Hook V\footnote{Alexander Blade, \url{http://www.dev-c.com/gtav/scripthookv}, 2018.} library, which enables access to metadata about the current scene.
Images and labels were captured with a resolution of \SI{1028x720}{\px} at a framerate of \SI{15}{fps}.
\Cref{tab:labels} shows an overview of the recorded data.
Note that for the experiments conducted for this work, we only made use of the depth maps, pixel-accurate masks for cars and pedestrians, the timestamp, as well as the weather conditions.
The numerous and various images and ground-truth annotations provided by \gls{dataset} are a precious data source for expanding the presented \gls{mt} approach.
They can furthermore serve as training data for entirely different tasks, such as video frame prediction.

\begin{figure}

  \begin{subfigure}{\linewidth}
    \centering
    \includegraphics{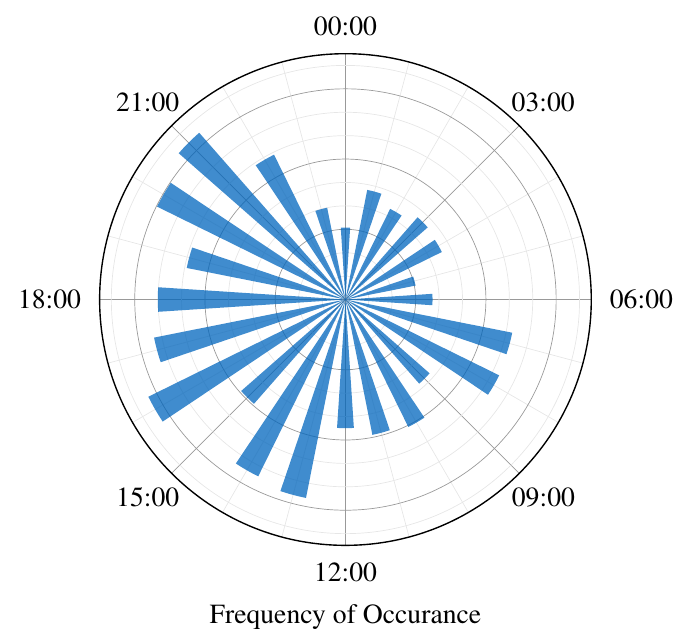}
    \subcaption{Time of Day}
  \end{subfigure}

  \vspace{0.5cm}

  \begin{subfigure}{\linewidth}
    \centering
    \includegraphics{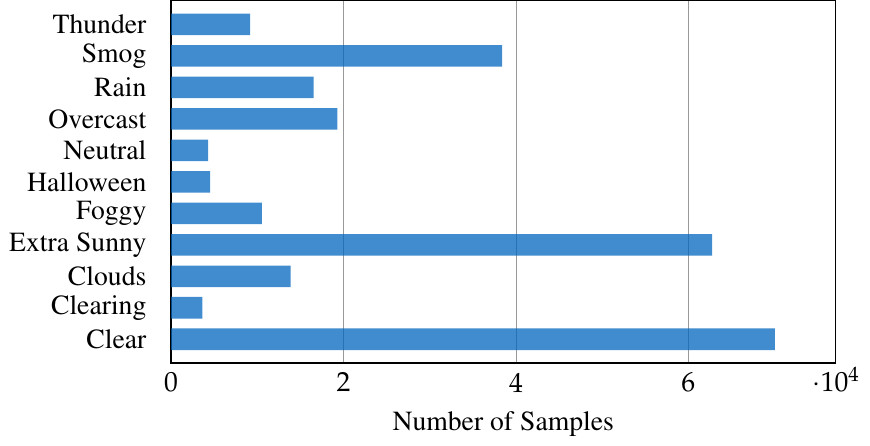}
    \subcaption{Weather Conditions}
  \end{subfigure}

  \caption{Distribution of samples over (a) time and (b) weather conditions in the training set of synMT.}
  \label{fig:dist_labels}

\end{figure}

In order to obtain disjoint subsets of samples for training and testing, the map area was split into non-overlapping areas.
A 2D-histogram over the position of all collected samples with a bin size of approximately \SI{65x65}{\square\meter} was computed to determine areas with available samples.
A test set was compiled by randomly selecting \num{100} bins from the dataset and choosing a subset of samples from them.
The test areas were buffered with a width of approximately \SI{65}{\meter} in order to clearly separate test and training areas.
Remaining samples, which overlap with neither the test areas nor their respective buffers, were kept for training.
The final subsets for training and testing consist of \num{2.5e5} and \num{e3} samples, respectively.
\Cref{fig:dist_labels} shows the distribution of the two auxiliary labels used for the experiments described in \cref{subsec:results} for the training set.
Note that the subset sampled from the test areas was designed to show similar statistical characteristics.

\begin{table*}[t]
  \caption{Performance on the synMT test set after optimization in different single- and \gls{mt} configurations.}
  \label{tab:results}

  \setlength{\tabcolsep}{0pt}

  \begin{tabularx}{\linewidth}{@{}S[table-format=2.2, table-auto-round, table-omit-exponent, fixed-exponent=-2, table-align-text-post=false]S[table-format=1.4, table-auto-round]S[table-format=3.0, table-auto-round]S[table-format=2.2, table-auto-round, table-omit-exponent, fixed-exponent=-2, table-align-text-post=false]@{}}

    \toprule
    \header{@{}X}{$\tau_1$ -- Semantic Segmentation} & \header{@{\hspace{5pt}}X}{$\tau_2$ -- Depth Estimation} & \header{@{\hspace{5pt}}X}{$\tau_3$ -- Time Estimation} & \header{@{\hspace{5pt}}X@{}}{$\tau_4$ -- Weather Classification} \\
    \header{@{}X}{MIoU} & \header{@{\hspace{5pt}}X}{RMSE of \functionAt[r]{d} (in \si{\meter})} & \header{@{\hspace{5pt}}X}{RMSCTD (in \si{\minute})} & \header{@{\hspace{5pt}}X@{}}{Accuracy} \\
    \cmidrule(r){1-1} \cmidrule(lr){2-2} \cmidrule(lr){3-3} \cmidrule(l){4-4}
    \best{74.28\%}      & \textemdash     &   \textemdash     &   \textemdash       \\  
    \textemdash         & 0.08257         &   \textemdash     &   \textemdash       \\  
    \textemdash         & \textemdash     &   \best{83}       &   \textemdash       \\  
    \textemdash         & \textemdash     &   \textemdash     &   \best{79.90\%}    \\  
    0.7283 \%           & 0.06709         &   \textemdash     &   \textemdash       \\  
    0.7185 \%           & 0.1132          &   435.6           &   \textemdash       \\  
    0.6558 \%           & \best{0.05704}  &   \textemdash     &   0.7015 \%         \\  
    0.7042 \%           & 0.1793          &   110.4           &   0.6982 \%         \\  

    \bottomrule

  \end{tabularx}
\end{table*}

\subsection{Network Architecture and Training}
\label{subsec:training}

In order to study the proposed approach, we chose the commonly used tasks \gls{side} and \gls{semseg} as our \glspl{main}.
By studying the clues on which the predictions for the \glspl{main} are based upon, tasks that encourage the network to learn features that may alleviate possible pitfalls can be found.
Methods for \gls{side} have been found to largely rely on image gradients as features \citep{Koch18}.
The time of day, especially different lighting conditions, distort them severely.
By adding a \emph{regression} for the current \emph{time of day} as an \gls{aux}, we push the network towards learning to differentiate between gradients caused by geometry and lighting.
Methods for \gls{semseg} need to clearly separate between actual objects that are expected to be labeled as such, and noise, \ia in the form of raindrops, snow, or mirror images.
Furthermore, objects may change appearance depending on the visibility conditions.
We, therefore, added the \emph{classification} of \emph{weather conditions} as a second \gls{aux} to force the network to explicitly learn features that are useful for identifying this influence.
Both tasks were optimized by single-task training in preliminary experiments and could be proven to converge to good results after moderate training time.

Since the \gls{e} part of the network primarily serves as a feature extractor, pre-training is feasible and reasonable.
State-of-the-art \gls{cnn} architectures for \gls{ir}, such as \glspl{resnet} \citep{He16}, are well suited as a basis for this part.
The goal of \gls{ir}, however, is to solve a global classification task.
Hence, a modification is necessary in order to obtain pixel-wise outputs as desired for \gls{semseg} and \gls{side}.
The concept of \glspl{fcn} \citep{Long15} enables transformation.
One instance of such architectures is \emph{DeepLabv3} \citep{Chen17}, which uses a \gls{resnet} as its feature extractor and employs \gls{aspp} \citep{Chen18b} to retain spatial information at different scales.
We utilized a re-implementation of this architecture with a pre-trained \gls{resnet}50 and replaced the final \num{1x1} \gls{conv} layer with task-specific \glspl{d}, as suggested by \citet{Kendall18}.
An overview of the architecture is shown in \cref{fig:overview}.

For each of the \gls{side} and \gls{semseg} \glspl{d}, we employ two $3 \times 3$ \gls{conv} layers with \gls{relu} activations, followed by a $1 \times 1$ convolution and bilinear interpolation.
The weather classification branch consists of a \gls{relu}-activated $5 \times 5$ \gls{conv} layer, two $3 \times 3$ \gls{max} layers with stride 3, and a $3 \times 3$ \gls{conv} layer with \gls{relu} activation between them, followed by a $1 \times 1$ convolution and a \gls{fc} layer with 11 neurons, corresponding to the number of classes for this label.
The time regression branch is identical to the weather classification branch except for the first \gls{max} layer, with a size of $5 \times 5$ and a stride of 5.
As a scalar output is expected, the final \gls{fc} layer features only one neuron.

For each atomic task $\tau_i \in \Tau$, an appropriate \gls{loss}, computed using the corresponding ground-truth label $y_{\tau}$, was selected or designed.
The commonly used combination of \gls{softmax} and \gls{ce} functions was employed for the pixel-wise and scalar classification tasks.
The pixel-wise and scalar regression tasks were assessed by modified \gls{mse} functions.
For the analysis of traffic scenes, the relative distance of closer objects is much more relevant than that of objects far away from the vehicle.
Therefore, we scaled the ground-truth depth non-linearly, such that instead of the absolute distance $d$ from the sensor, we estimated a scaled range
\begin{align}
  \functionAt[r]{d} &= 1 - \frac{\log(d)}{\log(1000)} \quad \text{.}
\end{align}
By applying this mapping, relevant depth values in the region of \SI{1}{\meter} to \SI{1}{\kilo\meter} were logarithmically scaled to $[0, 1]$.
We additionally clipped all values with $d < \SI{1}{\meter}$ and $d > \SI{1000}{\meter}$ to the minimum of $\functionAt[r]{d} = 0$ and maximum of $\functionAt[r]{d} = 1$, respectively.
By optimizing the \gls{mse} of $\functionAt[r]{d}$, the network is encouraged to predict closer ranges with high precision while tolerating large errors in further distances.

When calculating the difference between a predicted point in time $t'$ and the respective ground-truth $t$, each given as the minute of the day, the periodicity of the values has to be considered.
We calculated a \gls{sctd}
\begin{align}
  \begin{split}
    \functionAt[sctd]{t, t'} = \min \{ &(t - t')^2, (t - t' + 1440)^2, \\
                              & (t - t' - 1440)^2 \}
  \end{split}
\end{align}
and minimized the single-task loss
\begin{align}
  \functionAt[L_\text{time}]{\DataVec, \datalabel_{\text{time}}, \datalabel_{\text{time}}^\prime; \V{\omega}_\text{time}} &= \frac{\functionAt[sctd]{\datalabel_{\text{time}}, \datalabel'_{\text{time}}}}{b} \cdot 10^{-5}
\end{align}
given a \gls{batch} of size $b$.
Note that scaling the values by $10^{-5}$ brings them down to more common loss values and was solely done for convenience regarding the implementation.

Training was conducted using an implementation of the network architecture described in \cref{sec:approach} and \gls{dataset}, introduced in \cref{subsec:data}.
In each experiment, a different combination of tasks was considered.
Namely the four selected individual tasks as single-task reference ($\Tau = \{ \tau_1 \}, \{ \tau_2 \}, \{ \tau_3 \}, \{ \tau_4 \}$), the combination of all four ($\Tau = \{ \tau_1, \tau_2, \tau_3, \tau_4 \}$), the well-established combination of \gls{side} and \gls{semseg} ($\Tau = \{ \tau_1, \tau_2 \}$) as well as this traditional combination, enhanced by either weather classification or time regression ($\Tau = \{ \tau_1, \tau_2, \tau_3 \}$ and $\Tau = \{ \tau_1, \tau_2, \tau_4 \}$).

The loss weighting variables $c_\tau$ were initialized with a value of $\abs{\Tau}^{-1} = 0.25$ for all tasks $\tau$ in each experiment.
The spatial pyramid pooling rates were set to \num{1}, \num{2}, and \num{4}, yielding a final output stride of \num{16} before interpolation.
Since the deep \gls{mt} network in conjunction with large input images is demanding in GPU memory, a small batch size of $b = 4$ was utilized for optimization with a maximum \emph{learning rate} of $\alpha = 10^{-5}$ for the \emph{Adam} optimizer \citep{Kingma15}.

We performed optimization on an NVIDIA DGX-1v server, equipped with eight NVIDIA Tesla V100 (\SI{16}{\giga\byte}) GPUs, for approximately \num{12} days.
Using the NVIDIA DGX-1v \emph{(Volta)} over an NVIDIA DGX-1 \emph{(Pascal)} decreased the computation time per iteration by approximately 20\%.
Note that since no parallelization was implemented, each GPU was used for a different experiment.
Hence, optimization of the network could be conducted on a single NVIDIA Tesla V100 in roughly the same time, or an NVIDIA Tesla P100 and even NVIDIA Titan X (Pascal) with a minor increase in training time.

\subsection{Results and Discussion}
\label{subsec:results}

\begin{figure}

  \centering

  \includegraphics{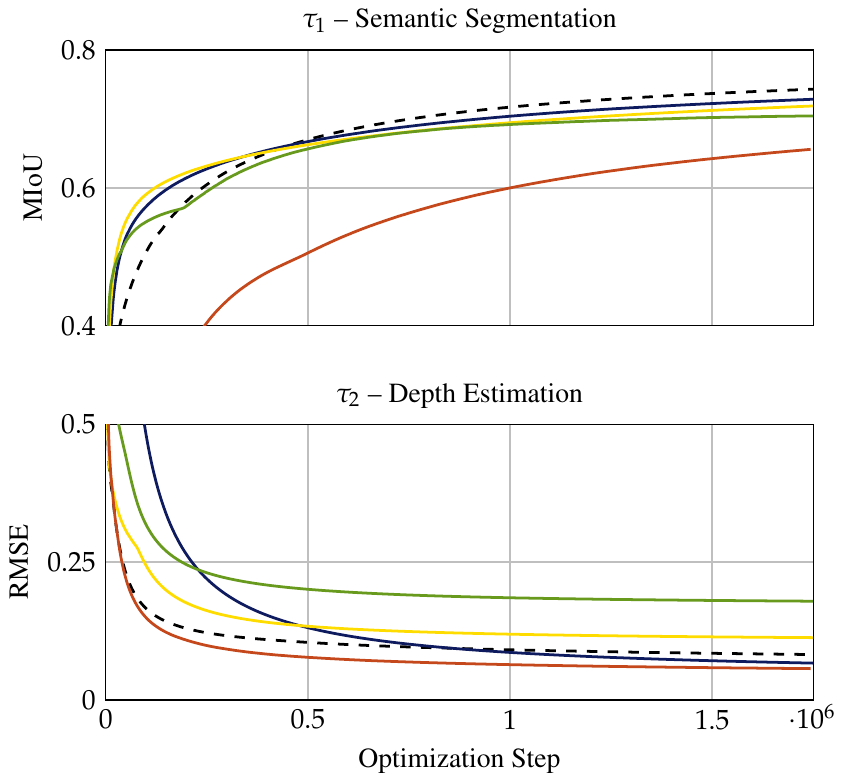}
  \includegraphics{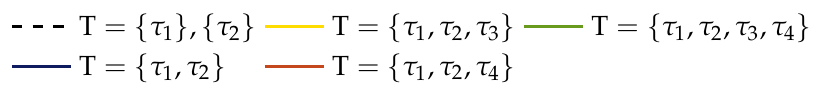}

  \caption{Performance for intermediate snapshots of the network parameters, showing the influence of \gls{mt} learning as a regularization  measure by yielding faster convergence. Optimization was conducted for \num{1.75e6} iterations (28 epochs) in 12 days on an NVIDIA DGX-1v GPU server.}
  \label{fig:plt_performance}

\end{figure}

\begin{figure}[t]

  \centering

  \begin{subfigure}{.75\linewidth}
    \fbox{\includegraphics[width=\linewidth]{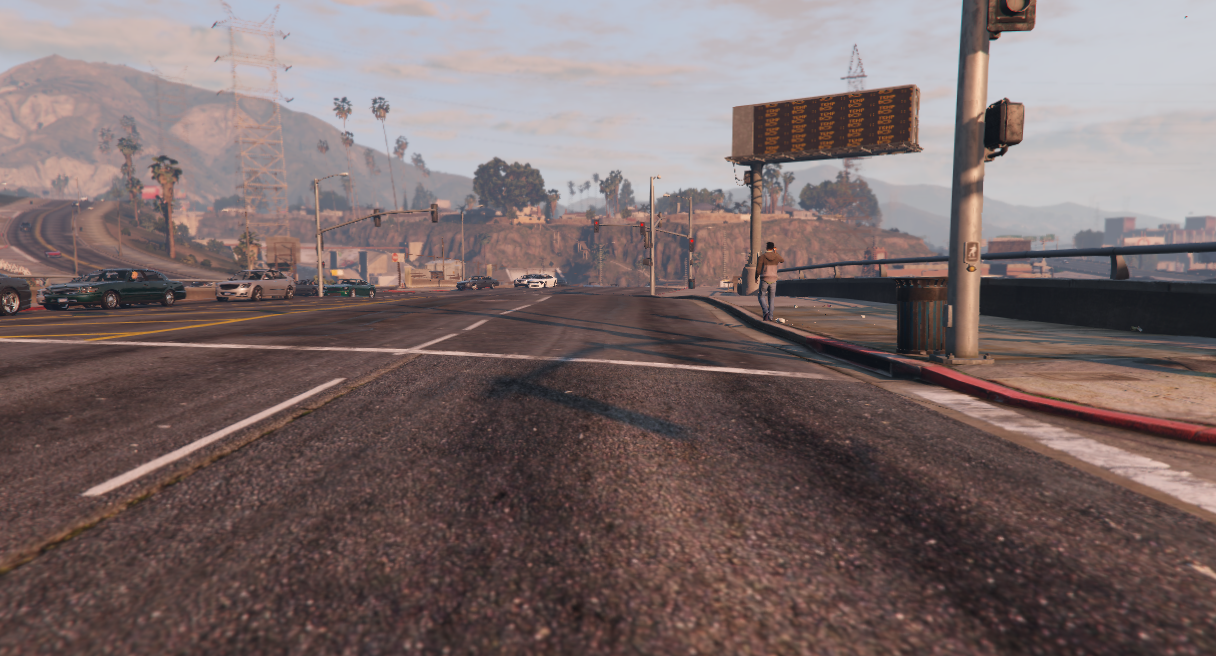}}
    \subcaption{RGB Input}
  \end{subfigure}

  \begin{subfigure}{\linewidth}
    \fbox{\includegraphics[width=.48\linewidth]{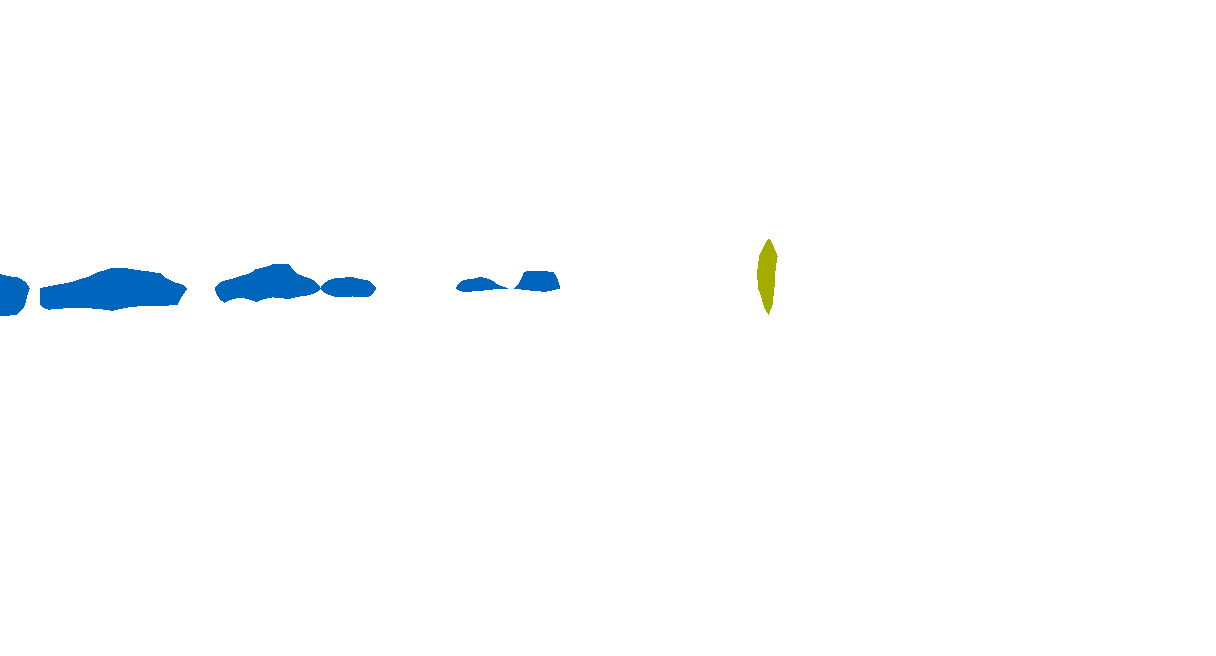}}
    \hfill
    \fbox{\includegraphics[width=.48\linewidth]{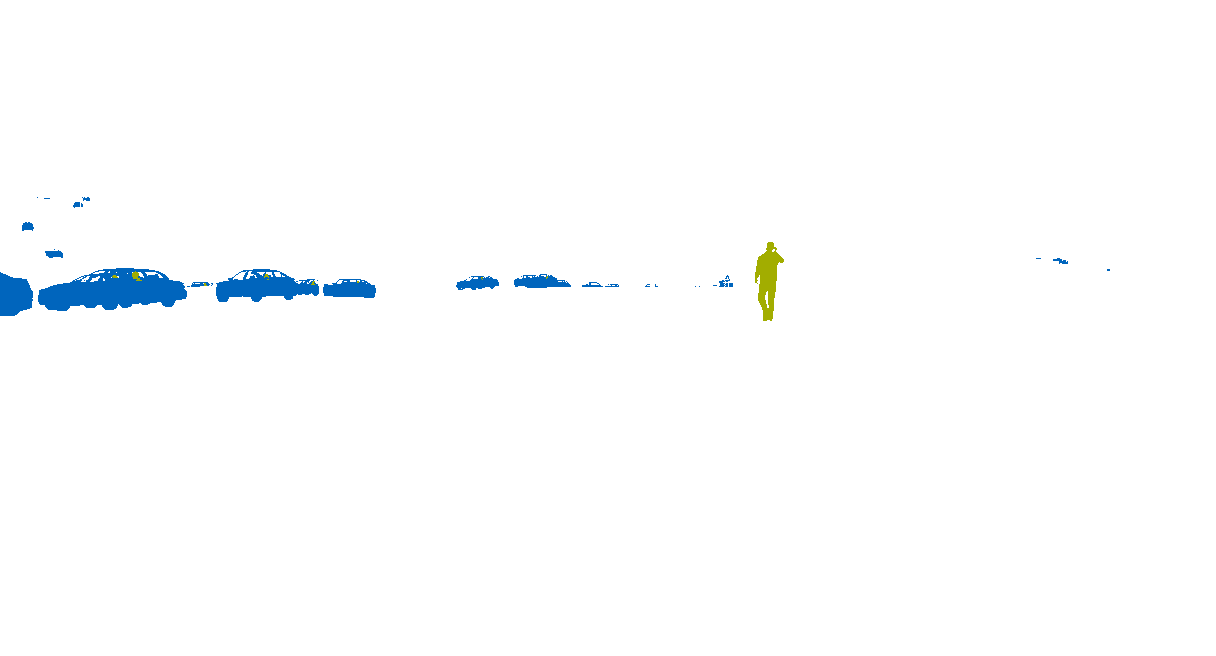}}
    \subcaption{Semantic Segmentation}
  \end{subfigure}

  \begin{subfigure}{\linewidth}
    \fbox{\includegraphics[width=.48\linewidth]{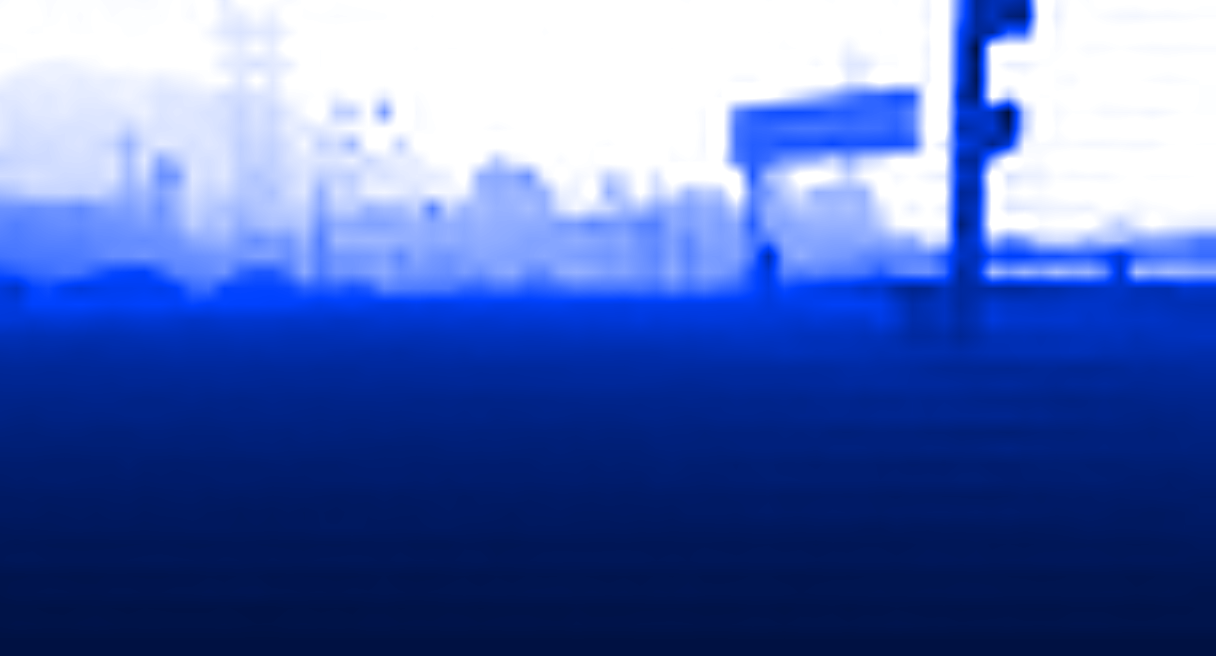}}
    \hfill
    \fbox{\includegraphics[width=.48\linewidth]{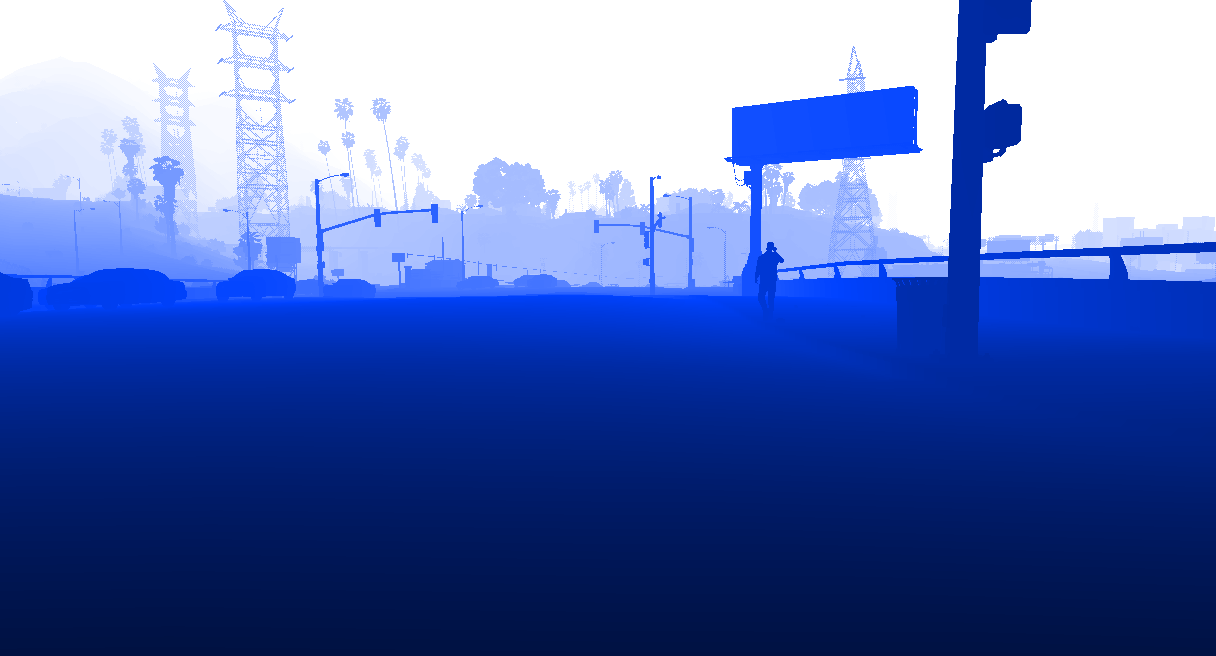}}
    \subcaption{Depth Estimation}
  \end{subfigure}

  \caption{Prediction results (left) and ground truth (right) for (b) \gls{semseg} and (c) depth estimation, derived from an RGB input image (a). The predictions illustrate the good performance of the network, despite the large output stride of 16.}
  \label{fig:results}

  \vspace{-0.25cm}

\end{figure}

The performance of the network was evaluated using the test set and an interpretable error metric for each task, \ie the \gls{rmse} for \gls{side}, \gls{miou} for \gls{semseg}, accuracy for weather prediction, and the \gls{rmsctd} for time regression.
A summary of the results is given in \cref{tab:results}, which shows the single-task baseline in the first part and the \gls{mt} results for different $\Tau$ in the second part.
As expected, the \glspl{aux} achieve high results overall with the best performance in the single-task settings.
Since they were solely added to support the network optimization during the training phase, the results are not further discussed here.
Note that the single notably higher value for $\tau_3$ in $\Tau = \{ \tau_1, \tau_2, \tau_3 \}$ was caused by a high weighting coefficient $c_{\tau_3}$ that diverged in the early training stages and could not be recovered from an unfavorable but apparently stable state.
In our experiments, we observed that the coefficients did not necessarily converge to similar values at each re-initialization, as reported by \citet{Kendall18}.

While \gls{semseg}, in general, did not profit from any \gls{mt} setup in our experiments, the best depth prediction was achieved when optimizing for $\Tau = \{ \tau_1, \tau_2, \tau_4 \}$, \ie adding weather classification to the \glspl{main}.
The results for $\Tau = \{ \tau_1, \tau_2, \tau_3, \tau_4 \}$, \ie both \glspl{main} and \glspl{aux}, suggest furthermore that it is possible to learn a common representation that generalizes to four different tasks.
The achieved moderate performance, however, reveals that $\Tau$, especially the \glspl{aux}, has to be chosen carefully.
Further experiments should be conducted in order to find \glspl{aux} that support the \glspl{main} desired for a selected application more effectively.
However, our first experiment showed that even seemingly unrelated \glspl{aux} can improve the results of \glspl{main}.

In order to study additional benefits of an extended \gls{mt} setup, we evaluated the network performance at intermediate optimization states.
\Cref{fig:plt_performance} shows the results of the \glspl{main} for each $\Tau$.
As expected, \glspl{aux} serve as regularization and thus facilitate faster convergence.
This became especially apparent for \gls{side}, where $\Tau = \{ \tau_1, \tau_2, \tau_3 \}, \{ \tau_1, \tau_2, \tau_4 \}$ converged considerably faster than the traditional \gls{mt} setup $\Tau = \{ \tau_1, \tau_2\}$.
For \gls{semseg}, the single-task setup yielded best results but converged slowly.

Samples from \gls{dataset} with predictions and gound-truth are shown in \cref{fig:results}.
The influence of the high output stride, which requires interpolation with a factor of \num{16}, is clearly visible in the pixel-wise outputs of the \glspl{main}.

\section{Conclusion}
\label{sec:conclusion}

We presented a new concept for improving \gls{mt} learning by adding seemingly unrelated \glspl{aux} to the set of learned tasks in order to improve the performance of the ultimately desired \glspl{main}.
\Glspl{aux} should be reasonably easy to learn and utilize annotations that require little effort to acquire.
Due to their properties, they can, thus, serve as a regularization  measure and improve performance, robustness, and training speed of the network.

To study our proposed approach, we applied the concept to \gls{vb} \gls{rsu} using our new synthetic \gls{mt} dataset \gls{dataset} and an extended state-of-the-art \gls{cnn} architecture.
The results showed that \glspl{aux} can indeed boost the performance of the \glspl{main}.
Several combinations of tasks were studied and the reported influence on the results can serve as a starting point for the evaluation of application-specific \gls{mt} systems.
Since the correlation between pairs and sets of main and \gls{aux} is expected to be alike in synthetic and real environments, drawn conclusions can be applied to the design of real-world data acquisition campaigns and networks.

\section*{Acknowledgements}
\label{sec:ack}

The work of Lukas Liebel was supported by the Federal Ministry of Transport and Digital Infrastructure (BMVI).
We gratefully acknowledge the computing resources provided by the Leibniz Supercomputing Centre (LRZ) of the Bavarian Academy of Sciences and Humanities (BAdW) and their excellent team, namely Yu Wang.
Finally, we thank our colleagues from the Computer Vision Research Group for providing precious feedback.

\bibliographystyle{plainnat}
\bibliography{bib_cleaned}
\end{document}